\documentclass[conference]{IEEEtran}
\IEEEoverridecommandlockouts
\usepackage{cite}
\usepackage{amsmath,amssymb,amsfonts}
\usepackage{algorithmic}
\usepackage{graphicx}
\usepackage{textcomp}
\usepackage{hyperref}
\usepackage{booktabs}
\usepackage{balance}
\usepackage{xcolor}
\usepackage{ulem}
\def\BibTeX{{\rm B\kern-.05em{\sc i\kern-.025em b}\kern-.08em
    T\kern-.1667em\lower.7ex\hbox{E}\kern-.125emX}}

\begin{document}

\title{ILDiff: Generate Transparent Animated Stickers \\ by Implicit Layout Distillation
}

\author{
% \name{
Ting Zhang$^{\dagger}$ \qquad
Zhiqiang Yuan$^{\dagger}$ \qquad Yeshuang Zhu \qquad Jie Zhou \qquad Jinchao Zhang$^{\star}$  \thanks{$^{\dagger}$ The first two authors contributes equally }
\thanks{$^{\star}$ Jinchao Zhang is the corresponding author }
% } 
% \address{ 
\\  Wechat AI, Tencent 
% }
}

% \and
% \and
% \and
% \and
% \and

\maketitle

\begin{abstract}
High-quality animated stickers usually contain transparent channels, which are often ignored by current video generation models.
To generate fine-grained animated transparency channels, existing methods can be roughly divided into video matting algorithms and diffusion-based algorithms.
The methods based on video matting have poor performance in dealing with semi-open areas in stickers, while diffusion-based methods are often used to model a single image, which will lead to local flicker when modeling animated stickers.
In this paper, we firstly propose an ILDiff method to generate
animated transparent  channels through implicit layout distillation, which solves the problems of semi-open area collapse and
no consideration of temporal information in existing methods.
Secondly, we create the Transparent Animated Sticker Dataset (TASD), which contains 0.32M high-quality samples with transparent channel, to provide data support for related fields.
Extensive experiments demonstrate that ILDiff can produce finer and smoother transparent channels compared to other methods such as Matting Anything and Layer Diffusion.
Our code and dataset will be released at link
\textcolor{blue}{\href{https://xiaoyuan1996.github.io}{https://xiaoyuan1996.github.io}}.
\end{abstract}


\section{Introduction}
% dehazing 方法 ASM

As a commonly used medium on social platforms, animated stickers play an important role in conversation and communication \cite{lei2016rating}\cite{huang2017overlapping}.
To obtain the available animated stickers, existing video generation algorithms can be directly applied \cite{zhou2024survey}\cite{zhang2023i2vgen}.
However, most of the results generated by these models are in the RGB color space and lack the transparent channel \cite{wu2023tune}\cite{du2024learning}, which plays a crucial role in high-quality stickers.

To predict a transparent channel based on an RGB image, the cutting-edge algorithms can be roughly divided into video matting-based algorithms \cite{li2024matting}\cite{ravi2024sam} and diffusion-based algorithms \cite{zhang2024transparent}\cite{wang2024diffusion}.
The video matting method usually trains the neural network based on a large amount of data to directly predict the transparent channel. 
Diffusion method, such as layer diffusion\cite{li2024matting}, encodes transparent channel into the potential manifold of the pre-trained diffusion model, and transform the model into a transparent image generator by slightly changing the pre-trained distribution.

Although the above two methods have made some progress in natural scenes, they inevitably encounter some problems when directly applied to the modeling of transparent animated stickers. 
Due to the singleness of the background color in stickers, the video matting-based method does not work well when processing semi-open areas, where the foreground and background colors are consistent.
While the current diffusion-based method is used to model a single image without considering the temporal information\cite{zhang2024transparent}\cite{wang2024diffusion}, which will cause local flicker when modeling animated stickers. 

In this paper, an ILDiff method is proposed to generate transparent animated channels through implicit layout distillation, which solves the problems of semi-open area collapse and no consideration of temporal information in existing methods.
Specifically, we manage to add implicit layout information via distill SAM to the current diffusion-based method, thus to characterize the semi-open area in the case of consistent foreground and background colors, which is difficult to be processed by the classical segmentation model. 
Next, we design a temporal modeling branch to endow ILDiff with the ability of temporal processing, thereby improving the local flickering problem encountered by traditional diffusion-based methods.
Lastly, we construct a transparent animated sticker dataset TASD, which contains 0.32M high-quality samples with transparent channel, to provide data support for this field.
% verify the proposed method, and 
The experiments qualitatively and quantitatively show that ILDiff can produce more refined and smoother transparent channels compared with other methods.
The main contributions of our work are as follows:
\begin{itemize}

\item We propose ILDiff, a method that leverages implicit layout distillation to generate animated transparent channels, which achieves a more refined and smoother result compared to other state-of-the-art methods. 

\item  A high-quality transparent animated sticker dataset TASD with corresponding benchmark is contributed, aiming to provide more data support for intelligent creation.

\end{itemize}

\vspace{-5pt}
\section{Methods}

\begin{figure*}[htbp]
	\centering
	\includegraphics[width=1\textwidth]{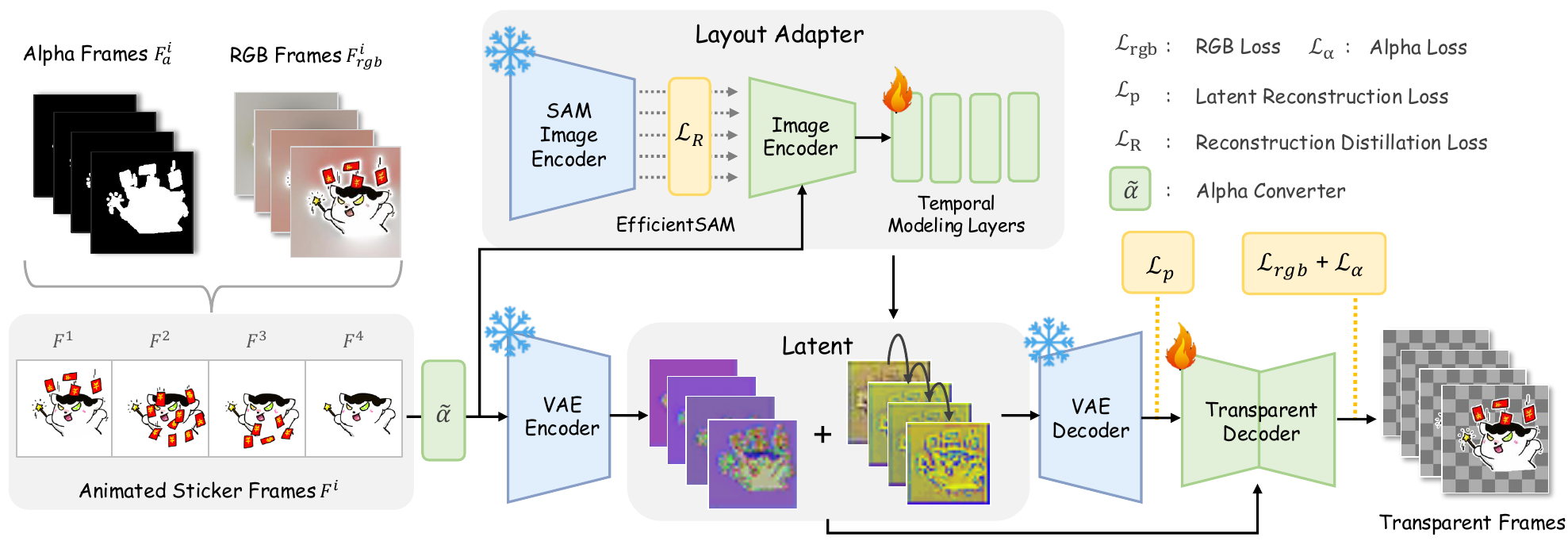}
	\caption{Framework of the proposed ILDiff model.
    Compared with layer diffusion, ILDiff adds a layout adapter, which learns the implicit layout information in animated stickers by distilling SAM and constructs a temporal modeling branch to improve the local flickering problem encountered by traditional diffusion-based methods.
    During training, the loss committee consisting of $\mathcal{L}_g$, $\mathcal{L}_{rgb}$, and $\mathcal{L}_p$ is used to jointly optimize the model.
 } 	\label{intro1} 
\vspace{-10pt}
\end{figure*}

\subsection{Diffusion Models for Transparent Image Generation}

Diffusion models \cite{sohl2015deep}\cite{song2020score} are a notable class of generative models that create new samples by gradually introducing noise to an initial data point and subsequently learning to reverse this process. Given a paired dataset $(\boldsymbol{x}, \boldsymbol{y}) \sim p_{\text{train}}(\boldsymbol{x}, \boldsymbol{y})$, where $\boldsymbol{x}$ represents a video clip consisting of $N$ frames, $\boldsymbol{x} = \{x^i | i = 1, 2, ..., N\}$, and $\boldsymbol{y}$ corresponds to the associated caption, these models aim to transform samples from a Gaussian distribution into the target distribution through iterative processes.
In the forward diffusion process, at each step $t$, a noisy image $x^i_t$ is generated using the formula $x^i_t = \sqrt{\alpha_t} x^i + \sqrt{1 - \alpha_t} \epsilon$, where $\epsilon \sim \mathcal{N}(0, \boldsymbol{I}_d)$ denotes Gaussian noise, $\boldsymbol{I}_d$ is the identity matrix of dimension $d$, and $\alpha_t$ determines the noise ratio at step $t$. The reverse process employs a learnable network $\mathcal{G}_{\theta_1}(x^i_t, t)$ to predict the noise and recover the clean image from the noisy input $x^i_t$. After training, the model begins with pure Gaussian noise $x_T \sim \mathcal{N}(0, \boldsymbol{I}_d)$ and iteratively refines this noise over $T$ steps to sample a clean image.
% Latent Diffusion Models (LDMs) \cite{rombach2022high} operate within a latent space makes the diffsusion process more efficient and computationally less intensive. 
% % In this approach, the latent representation of an image at time $t = 0$, denoted as $z_0$, undergoes a series of transformations into noisier latent representations $z_t$ following the distribution $q(z_t | z_{t-1}) = \mathcal{N}(z_t; \sqrt{1 - \beta_t} z_{t-1}, \beta_t \mathbf{I})$, where $\beta_t$ is a variance schedule parameter.
% Stable Diffusion (SD) is an exemplary method within this framework, combining the Variational Autoencoder (VAE) for efficient image compression and generation. The VAE architecture includes an encoder $\mathcal{E}_{sd}$ that maps RGB images to a latent space, and a decoder $\mathcal{D}_{sd}$ that reconstructs images from these latent representations.

% \vspace{-10pt}

% Early image diffusion methods typically relied on pixel-level data for training \cite{song2020denoising, kong2021fast, san2021noise}. 
Latent diffusion models \cite{rombach2022high} introduce the latent space, making the diffusion process more efficient and computationally less intensive. 
Among these, the classic Stable Diffusion (SD) combines a Variational Autoencoder (VAE) \cite{pinheiro2021variational} for diverse and stable image generation. 
VAE is trained by maximizing the likelihood function and its variational inference. Structurally, it includes an encoder $\mathcal{E}_{sd}$ that maps RGB images to a latent space, and a decoder $\mathcal{D}_{sd}$ that reconstructs images from these latent representations. 
% Furthermore,
Based on the latent diffusion, Layer Diffusion \cite{zhang2024transparent} extends the capabilities of large-scale pre-trained latent diffusion models to directly generate images with transparency. It trains two independent neural network models: a latent transparency encoder $\mathcal{E}_{tr}$ and a decoder $\mathcal{D}_{tr}$. 
During training, the encoder $\mathcal{E}_{tr}$ is responsible for transforming the RGB channels $I_r \in \mathbb{R}^{h \times w \times 3}$ and the transparent channel $I_\alpha \in \mathbb{R}^{h \times w \times 1}$ into an offset $\Delta z \in \mathcal{Z}$ in the latent space. This offset is added to the original latent vector $z \in \mathcal{Z}$ to form an adjusted latent vector $\hat{z}_{adj} = z + \Delta z$. The decoder $\mathcal{D}_{tr}$ then reconstructs the transparent image, yielding reconstructed RGB $\hat{I}_r$ and transparent channels $\hat{I}_\alpha$.
Layer Diffusion trains $\mathcal{E}_{tr}$ and $\mathcal{D}_{tr}$ using reconstruction losses $\|I_r - \hat{I}_r\|_2^2$ and $\|I_\alpha - \hat{I}_\alpha\|_2^2$ to ensure the quality of the decoded transparent channel. Additionally, a harmlessness measure is designed by comparing the image differences between the inputs and outputs of frozen encoder $\mathcal{E}_{sd}$ and decoder $\mathcal{D}_{sd}$,
making the latent distribution of the pre-trained model not be disrupted by introduced transparency channel.

\subsection{Implicit Layout Distill for Animated Transparent Channels}

% Due to the singleness of background color in stickers, when there is a semi-open area in the foreground that is the same color as the background, the existing video matting method often has unsatisfactory results.
Although layer diffusion can generate transparent channels for single images, there are two problems when dealing with animated stickers:
(a) Failure to consider timing information, which will cause local flicker when modeling animated stickers.
(b) As with the video matting method, the prior perception of semi-open areas needs to be injected.
% \textcolor{red}{Animated stickers often display bold colors, straightforward lines, and abstract actions. Most animated stickers include a main cartoon character surrounded by multiple accessories to create a expressive visual effect. However, this multi-component nature and broad, sparse movements pose challenges for the precise extraction of transparent channels in animated stickers.} 
% To address this, we propose a layout adapter module,  
% combining the transparent channel generation capability of layer diffusion with the layout extraction advantages of segmentation algorithms. 
To improve the above problems, we come up with a layout adapter module, which attempts to introduce layout priors while performing temporal modeling.
The layout adapter module consists of an image encoder and a temporal modeling layer. The image encoder captures the layout information of each component, while the temporal modeling layer integrates the inter-frame temporal information. The output feature of the layout adapter module is then incorporated into the latent space to guide the training process. 

For layout prior, what we need is an implicit feature, which can represent the semi-open area and is different from the strict segmentation information.
For this reason, we distill the Segment Anything Model (SAM) \cite{kirillov2023segment} to train the image encoder, making it better suited to the lightweight pipeline of the layout adapter. Followed by EfficientSAM \cite{xiong2024efficientsam}, we use mask image modeling to learn the high-level feature embeddings generated by the ViT-H \cite{dosovitskiy2020image} encoder of SAM, achieving effective feature representation.
Correspondingly, the reconstruction distillation loss function can be expressed as
$\mathcal{L}_{R} = \| f_{\text{sam}}(x) - f_{h}(x) \|^2_2$, 
where $\| \cdot \|^2_2$ denotes the $L_2$ norm distance,  $f_{\text{sam}}(x)$ and $f_{h}(x)$ represent the features output by the SAM image encoder and the features of the distilled lightweight image encoder in specific frame after MAE \cite{he2022masked} reconstruction and linear projection, respectively. 

% For temporal modeling
Furthermore, the temporal modeling layer effectively captures the temporal dependencies of each frame feature output by the distilled image encoder through a series of 3D convolutional layers. Each layer employs Group Normalization and ReLU activation functions to ensure stable and efficient training. We found that simple 3D convolutional layers can effectively achieve smooth transitions between frames of animated sticker component elements. The final output features are dimensionally transformed using a 2D convolution and adaptive average pooling to map to the same dimension in the latent space.

Finally, we fine-tune the layer diffusion's transparent decoder $\mathcal{D}_{tr}$ to decode the alpha channels from the latent space enriched with layout and temporal information. Similarly, the reconstruction quality is evaluated using the $L_2$ norm. $\mathcal{L}_{\alpha}$ and $\mathcal{L}_{rgb}$ represent the reconstruction losses for the alpha channel and color given by the $\mathcal{D}_{tr}$, respectively. Specifically, $\mathcal{L}_{\alpha} = \| F^i_{\alpha} - \hat{F}^i_{\alpha} \|^2_2$ and $\mathcal{L}_{rgb} = \| F^i_{rgb} - \hat{F}^i_{rgb} \|^2_2$, where \(\hat{F^i_*}\) denotes the reconstructed features of the $i$-th frame.
Besides, the reconstruction loss of the frozen VAE encoder-decoder $\mathcal{L}_{p} = \| \mathcal{E}_{sd}(F^i_p) - \mathcal{D}_{sd}(F^i_p) \|^2_2 = \| F^i_{p} - \hat{F}^i_{p} \|^2_2$ is also included to ensure the harmless injection of the layout adapter's output features into the latent space.

% The layout adapter module includes a image encoder and a temporal module, incorporating the layout information of each component and the inter-frame temporal information into the latent space to guide the training process. 
% Animated stickers are characterized by their lower frame rate and abstract semantics. Most animated stickers consist of many small and diverse animated elements rather than a single primary subject, achieving rich and expressive visual effects. 
% To enhance the model's ability to generate transparent channels for each element, we designed a layout adapter module. The layout adapter module consists of a \textit{segmentation feature extractor} and a \textit{temporal unit}, responsible for extracting layout information and inter-frame temporal information to guide the generation of refined transparent channels. 
% \textbf{a. Segmentation feature extractor.} In order to globally locate each structural element of the animated stickers, we distilled the Segment Anything Model (SAM) \cite{kirillov2023segment} image encoder to build the segmentation feature extractor. This approach retains SAM's spatial information extraction capability while meeting the computational efficiency requirements, thus better integrating into the layout adapter's processing pipeline. During the distillation process, we adopted Efficient SAM 

\section{Transparent Animated Sticker Dataset}

\begin{figure}[h]
\centering
\includegraphics [width=3.4in]{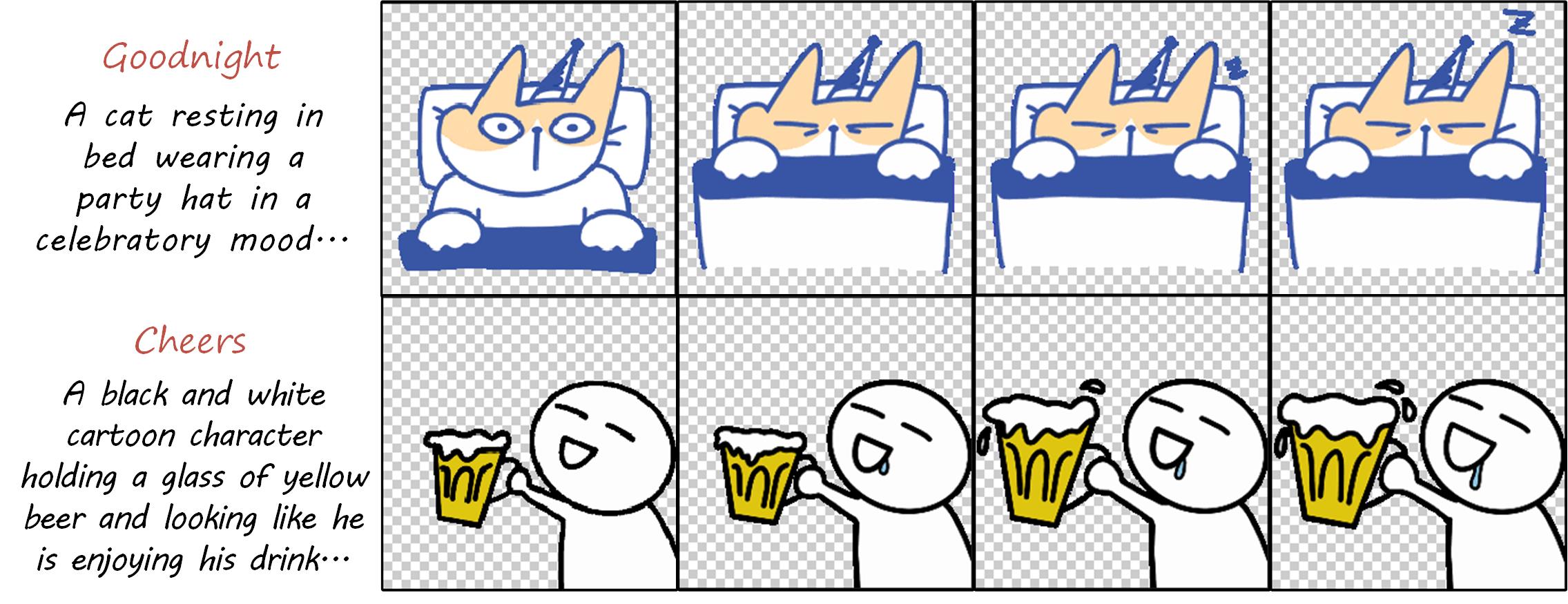}
\vspace{-10pt}
\caption{
Two samples of TASD, in which GIFs is framed for visualization.
The red word shows the trigger word.
% More results can be seen in \textcolor{blue}{\href{https://anonymous0722.github.io}{https://anonymous0722.github.io}}.
See \textcolor{blue}{\href{https://xiaoyuan1996.github.io}{here}} for animated samples.
}
\label{dataset_show}
\vspace{-7pt}
\end{figure}

\begin{figure}[!t]
% \vspace{-10pt}
\centering
\includegraphics [width=3.5in]{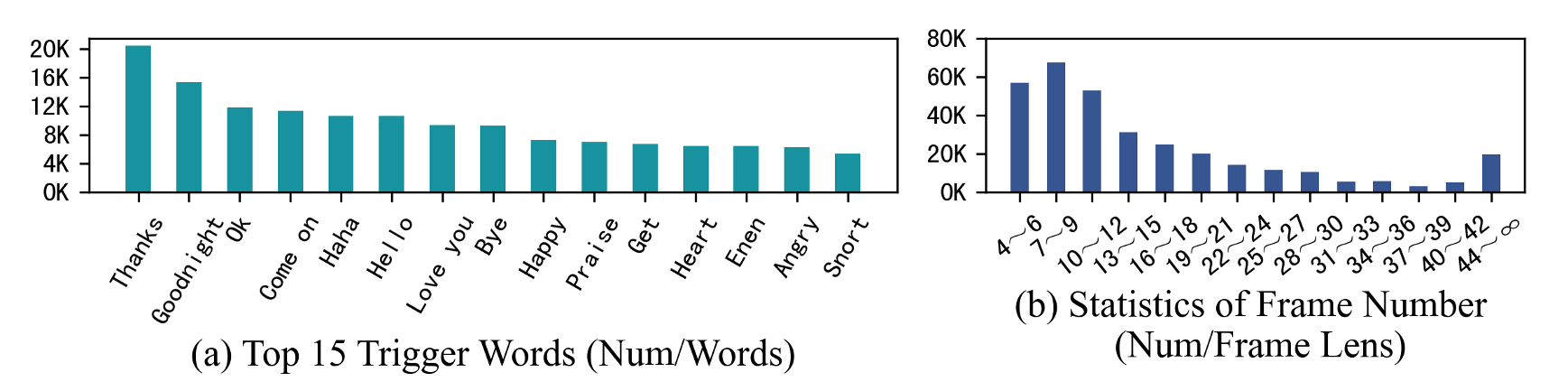}
\vspace{-20pt}
\caption{
Visual analysis of TASD.
(a) Frequency count of top 15 trigger words. 
(b) Statistics of frame number.
}
\label{data_freq}
\vspace{-10pt}
\end{figure}

% \subsection{Data Construction}
We collected 0.32M high-quality animated sticker data with transparent channels from private sources by retrieving 115 crawled words.
To obtain relevant text descriptions, we first manually labeled part of the data for the annotation task, which enabled us to fine-tune the caption model.
Next, the manually annotated data is used to perform supervised fine-tuning on VideoLlama to improve the accuracy of automatic annotation, so that the trained model can be directly used for visual caption generation.
In addition to the English version, we also provide a Chinese version to facilitate the needs of researchers.
% In addition to the description, 
Crawled words related to the visual samples are aggregated and used as trigger words.
Two examples of the TASD dataset are shown in Fig.\ref{dataset_show}, in which each sample contains a transparent animated sticker, a description, and a set of trigger words.
% \subsection{Data Statistics}
Table \ref{numerical_prop} shows some numerical properties of TASD, and the distribution of static attributes such as trigger words and frame number is shown in Fig.\ref{data_freq}.

To facilitate researchers' utilization, we manually selected 200 samples corresponding to the trigger words as the test set and named it TASD-T.
Subsequent experiments on TASD-T testset are utilized to evaluate the different model performance.

% \vspace{-10pt}
\section{Related Works}

\noindent
\textbf{Animated Sticker Generation}
% \textcolor{red}{
% Stickers, as a common communication medium on social platforms, have a great influence on the dialogue experience.
% Knn-Diffusion\cite{sheynin2022knn} has explored the generation of static stickers and achieved the generation of out-of-distribution stickers by using large-scale retrieval methods.
% Text-to-Sticker\cite{sinha2023text} first fine-tuned Emu\cite{dai2023emu} using millions of sticker images collected with weak supervision to induce diversity, and then proposed style tailoring fine-tuning method, which improves the visual quality of sticker generation by jointly fitting content and style distribution.
% To generate multi-frame stickers, Animated Stickers \cite{yan2024animated} used a two-stage fine-tuning process: first using weak in-domain data and then utilizing a human-machine loop strategy, which can effectively improve the motion quality.
% % VSD2M provided sticker data to the researcher and constructed a set of baselines to facilitate works related to animated stickers.
% In this paper, animated sticker generation is used as an application of video generation to validate the effectiveness of the proposed method.
% }
Stickers are a common medium for communication on social platforms, introducing a dynamic and engaging element to multimedia interactions. Knn-Diffusion \cite{sheynin2022knn} investigated static sticker generation, achieving the creation of out-of-distribution stickers using extensive retrieval methods. Text-to-Sticker \cite{sinha2023text} initially fine-tuned Emu \cite{dai2023emu} with millions of weakly supervised sticker images to promote diversity. Subsequently, they introduced a style-tailoring fine-tuning approach, enhancing the visual quality of stickers by simultaneously fitting content and style distributions. For multi-frame sticker generation, Animated Stickers \cite{yan2024animated} utilized a two-phase fine-tuning process: leveraging weak in-domain data at first, and then employing a human-machine loop strategy, which significantly improved motion quality. This paper improves animated sticker generation by generating transparent channel, achieving better visual effects across different application scenarios.

\begin{table}[!t]
 % \small
 % \vspace{-10pt}
  \caption{Numerical properties of the TASD dataset.}
\resizebox{\linewidth}{!}{
\begin{tabular}{ccc}
\hline
\textbf{\#Sample}    &  \textbf{Ave Description Length}   &   \textbf{Ave Frames Numbers}       \\ \hline
        0.32M   &       96.25      &                  16.71      \\ \hline
\textbf{\#Trigger Words} & \textbf{Annotation language}  & \textbf{Keyframe Index} \\ \hline
        115      &   En, Cn         &  \checkmark                    \\ \hline
\end{tabular}}
\vspace{-10pt}
\label{numerical_prop}
\end{table}

\noindent
\textbf{Transparent channel generation}
For transparent sticker generation, current research can be categorized into post-processing methods \cite{li2024vmformer}\cite{li2023videomatt} and generative approaches\cite{zhang2024transparent}\cite{wang2024diffusion}. Post-processing methods primarily focus on image segmentation and matting. These methods divide RGB images into different regions based on color, texture, and other visual features. Kirillov $et\ al.$ introduced the Segment Anything Model (SAM) \cite{kirillov2023segment}, a foundational model in computer vision capable of segmenting any object based on user prompts. Further, SAM was extended to SAM-2\cite{ravi2024sam}, which incorporates a Transformer structure with streaming storage and memory mechanisms to achieve coherent segmentation predictions in video sequences. Matting Anything \cite{li2024matting} builds on SAM's feature maps and mask outputs, adding a lightweight Mask-to-Matte module to predict the transparent channel. 
% However, these advanced post-processing methods still rely on some degree of user input to ensure the transparent channel's quality.
On the other hand, generative methods such as layer diffusion \cite{zhang2024transparent} present the concept of ``latent transparency''. 
This method encodes transparent channel transparency within the latent space of pre-trained latent diffusion models, such as Stable Diffusion, using separate encoder and decoder networks to transform and reconstruct transparency information. 
% This enables the generation of images with single or multiple transparent layers. 
% Our method targets the generation of animated stickers with transparent channels, considering the richer layout elements and significant variations between frames. 
Building on layer diffusion, we have extended its capability to handle animated images, and integrated the advantages of SAM's post-processing methods to inject implicit layout information.
% thereby achieving automated Transparent Animated Stickers.

\section{Experiments}

\subsection{Experiment Settings and Evaluation Metrics}

\noindent
\textbf{Implementation details.}
We use layer diffusion as the backbone and adds the layout adapter for training.
% To better distill SAM, we adopt the learning strategy of EfficientSAM.
During training, we fix the weights of VAE and SAM encoder and only fine-tune the layout adapter.
The depth of temporal layer is 5, which is selected through ablation experiments in Section \ref{exp}.

\noindent
\textbf{Comparison methods.}
We compare our model against current state-of-the-art transparent channel generation algorithms based on video matting (Matting Anything), segmentation (SAM, SAM2), and diffusion (Layer Diffusion).
For comparison methods, we utilize the provided pre-trained models for fair comparison.

\noindent
\textbf{Metrics.}
Peak signal to noise (PSNR) and structural similarity index (SSIM) are used to evaluate the effectiveness of various methods.
In addition, we introduce manual comparison to perceptually compare these methods in frame smooth and hole residue.

% \subsection{Ablative Study}

\subsection{Quantitative \& Qualitative Results}
\label{exp}

\vspace{-10pt}

\begin{table}[h]
    \centering
    \caption{Overall Average PSNR and SSIM for Different Methods.
    \textbf{Bold} and \uline{underline} indicate the best and second-best, respectively.}
    \label{tab:results}
\resizebox{\linewidth}{!}{
\begin{tabular}{cccccc}
\toprule
\textbf{Method} & \textbf{Matting Anything} & \textbf{SAM}    & \textbf{SAM-2}  & \textbf{Layer Diffusion} & \textbf{ILDiff}          \\ \hline
\textbf{PSNR}   & 24.6            & 22.86  & 18.05  & \uline{26.42}          & \textbf{28.04}  \\
\textbf{SSIM}   & 0.954          & 0.949 & 0.898 & \uline{0.966}         & \textbf{0.981} \\ 
\bottomrule
\end{tabular}
}
\vspace{-5pt}
\end{table}

\noindent
\textbf{Automatic Metrics.}
For method comparison, since SAM and Matting Anything are designed for single image processing, we segmented each frame of the expressions individually. We marked the center point and the outer rectangle of the ground truth alpha image to simulate user input of point prompts and box prompts. For SAM2, which has the capability of video segmentation, we provided prompts only for a single frame containing the subject. The comparison of PSNR and SSIM indicators of different methods is shown in Table \ref{tab:results}.
Our method is significantly ahead of other methods in both two indicators, which verifies that ILDiff can better model transparent channels with the help of implicit layout.
Among other comparison methods, Layer Diffusion ranks second in both indicators, followed by Matting Anything.

\begin{figure}[htbp]
\centering
\includegraphics [width=2.8in]{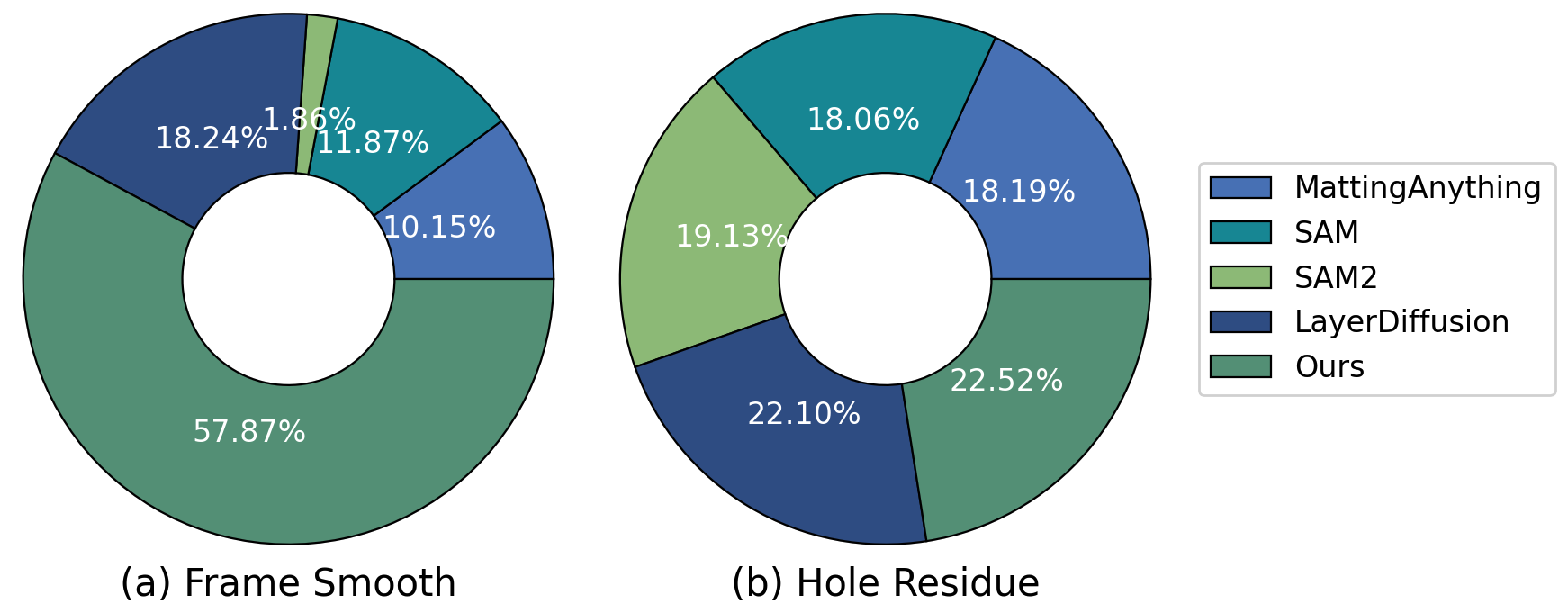}
\vspace{-10pt}
\caption{Manual comparison of generated transparent channel by different methods on (a) frame smooth and (b) hole residue.}
\label{manual_comp}
% \vspace{-7pt}
\end{figure}

% \begin{table}[h!]
%     \centering
%     \caption{Overall Average PSNR and SSIM for Different Methods}
%     \label{tab:results}
%     \begin{tabular}{lcc}
%         \toprule
%         \textbf{Method} & \textbf{Overall Average PSNR $\uparrow$} & \textbf{Overall Average SSIM $\uparrow$} \\
%         \midrule
%         ILDiff & 28.04 & 0.9807 \\
%         LayerDiffusion & 26.42 & 0.9658 \\
%         SAM1 & 22.86 & 0.9487 \\
%         SAM2 & 18.05 & 0.8982 \\
%         Matting & 24.60 & 0.9544 \\
%         \bottomrule
%     \end{tabular}
% \end{table}

\noindent
\textbf{User Preference.}
In addition to automatic metrics, we also introduce manual evaluation to overcome the limitation of existing metrics.
We request ten labelers to select the best generated stickers in terms of frame smooth and hole residue, and the related results is shown in Fig.\ref{manual_comp}(a) and Fig.\ref{manual_comp}(b), respectively.
In subjective evaluation, our method significantly outperforms other methods in terms of frame smoothness, indicating that our model can better model temporal relationships.
In terms of hole residue, our method is ahead of LayerDiffusion and can generate transparent channels that are closest to the ground truth.

\begin{figure}[h]
\vspace{-10pt}
\centering
\includegraphics [width=3.5in]{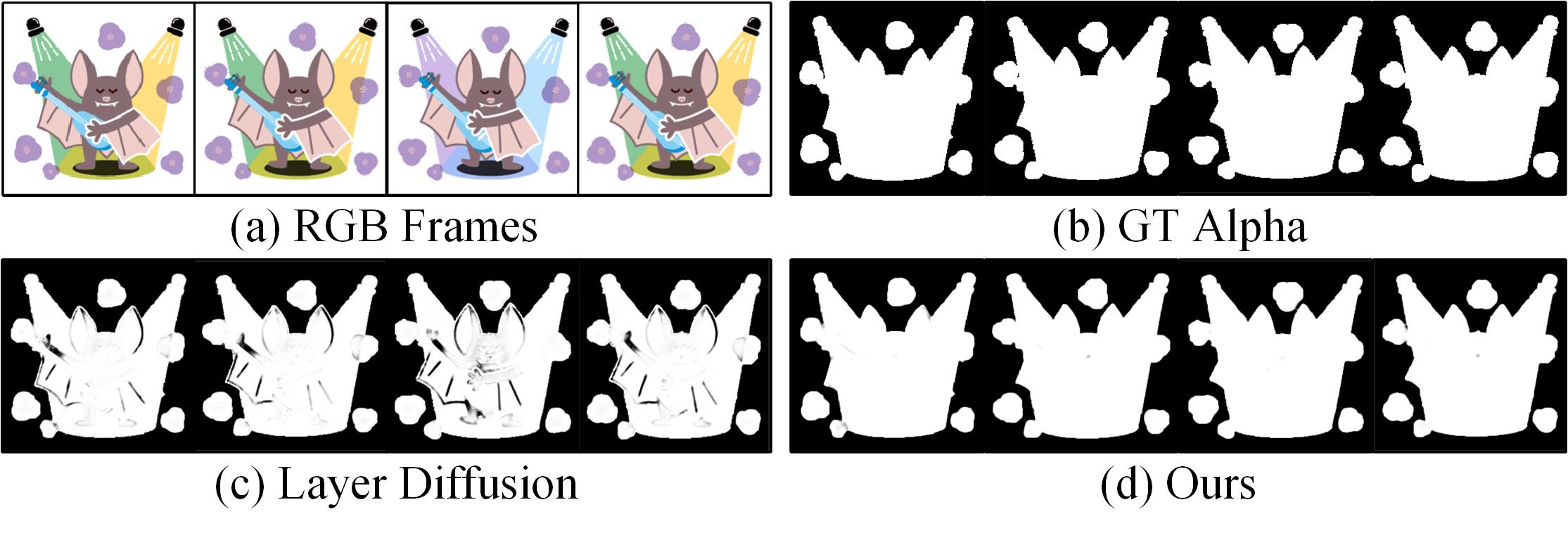}
\vspace{-20pt}
\caption{
Visual comparison for transparent channel generation between Layer Diffusion and Ours.
See \textcolor{blue}{\href{https://xiaoyuan1996.github.io}{here}} for more results.
}
\label{visual_1}
\vspace{-7pt}
\end{figure}

\begin{figure}[h]
\centering
\includegraphics [width=3.5in]{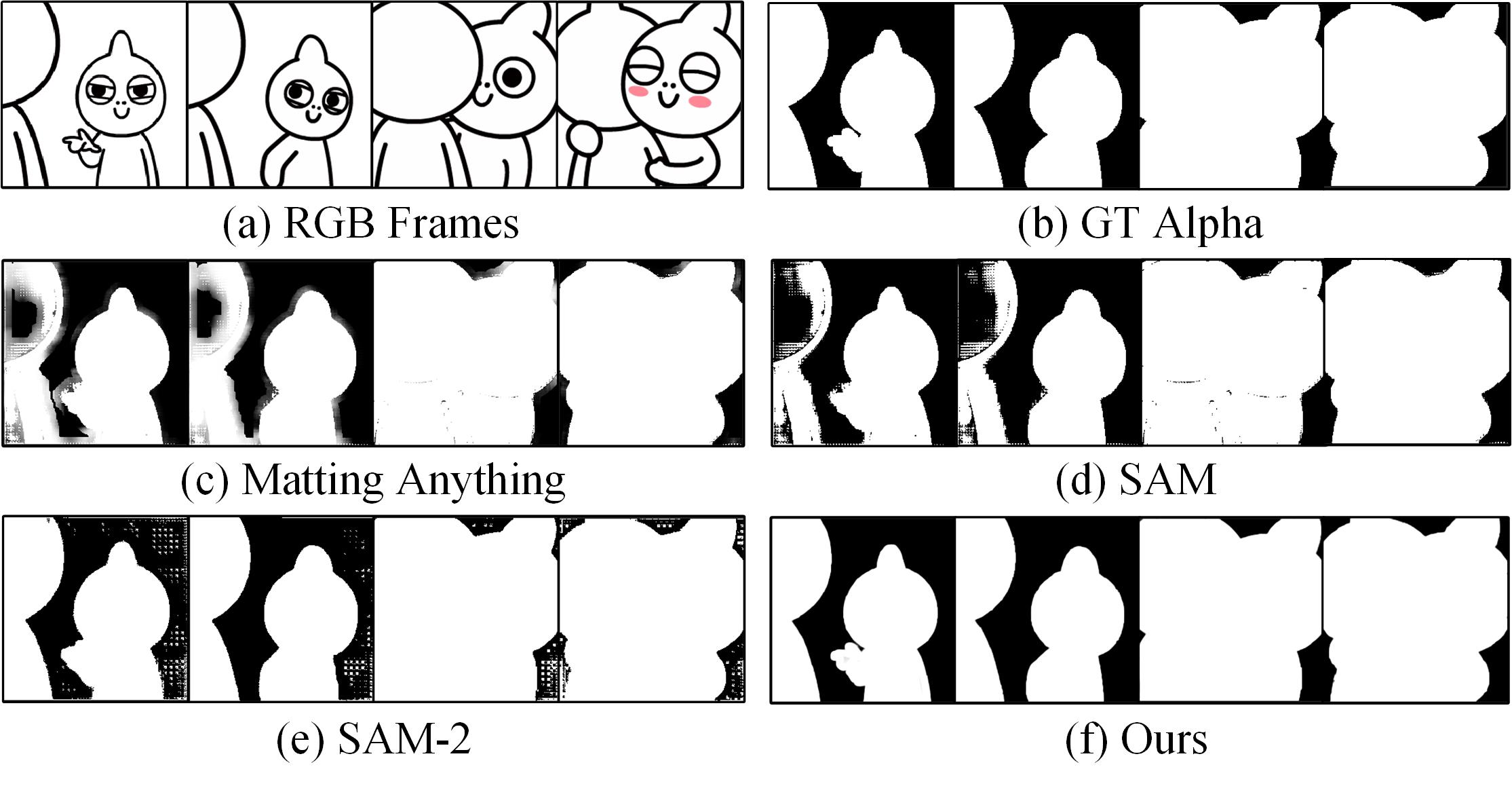}
\vspace{-20pt}
\caption{
Visual comparison for transparent channel generation between Matting Anything, SAM, SAM-2, and Ours.
See \textcolor{blue}{\href{https://xiaoyuan1996.github.io}{here}} for more results.
}
\label{visual_2}
\vspace{-7pt}
\end{figure}

\noindent
\textbf{Visual Comparison.}
Fig. \ref{visual_1} and Fig. \ref{visual_2} show the visual comparison of generated transparent channel by diffusion methods.
In Fig. \ref{visual_1}, compared to layer diffusion, our method can obtain a smoother transparent channel, which greatly alleviates the flickering problem caused by the former method.
In Fig. \ref{visual_2}, matting anything and SAM show a hole residue when dealing with the semi-open region, which is better avoided by our method due to the introduction of implicit layout.
In addition, compared with SAM-2, our method produces cleaner results in the background area and is closer to the ground truth of transparent channel.

\begin{table}[h]
\vspace{-5pt}
    \centering
    \caption{Ablation for depth of temporal layer in layout adapter.
    \textbf{Bold} and \uline{underline} indicate the best and second-best, respectively.}
    \label{tab:aba}
% \resizebox{\linewidth}{!}{
\begin{tabular}{ccccc}
\toprule
\textbf{\#Layer Depth} & \textbf{0} & \textbf{3}    & \textbf{5}  & \textbf{8}     \\ \hline
\textbf{PSNR}   & \uline{27.83}            & 27.81  &  \textbf{28.04}  & \uline{27.83}            \\
\textbf{SSIM}   & 0.978          & 0.978 & \textbf{0.981} & \uline{ 0.979}       \\ 
\bottomrule
\end{tabular}
% }
\end{table}

\noindent
\textbf{Ablative Study.}
As shown in Table \ref{tab:aba}, we ablated the depth of the temporal layer in the layout adapter to study the ILDiff setting that achieves the best generation performance.
When the depth of temporal layers is 0, ILDiff only adds implicit layout information compared to layer diffusion, which achieves suboptimal PSNR indicator.
When the layer depth is gradually increased to 5, the PSNR and SSIM indicators of the ILDiff model reach the optimal.
As the layer depth increases further, the quantitative performance of ILDiff begins to decay.

\section{Conclusion}

In this paper, we propose an ILDiff method to generate transparent animated stickers, which achieves a smoother transparent channel with less hole residue than other methods.
In addition, we also provide a high-quality transparent animated sticker dataset TASD, which aims to provide richer data resources for the field of intelligent creation.

\vspace{12pt}
\color{red}

\end{document}